\documentclass[lettersize,journal]{IEEEtran}
\usepackage{amsmath,amsfonts}
\usepackage{algorithmic}
\usepackage{algorithm}
\usepackage{array}
\usepackage[caption=false,font=normalsize,labelfont=sf,textfont=sf]{subfig}
\usepackage{textcomp}
\usepackage{stfloats}
\usepackage{url}
\usepackage{verbatim}
\usepackage{graphicx}
\usepackage{cite}
\usepackage{booktabs}
\usepackage{makecell}
\usepackage{multirow}
\usepackage{todonotes}
\usepackage{hyperref}
\usepackage{graphicx}
\usepackage{amsmath}
\usepackage{amsthm}
\usepackage{booktabs}
\usepackage[switch]{lineno}
\usepackage{algorithm}
\usepackage{algorithmic}
\usepackage{tabularx}
\usepackage{makecell}
\usepackage{xcolor}
\usepackage{amssymb}
\usepackage{setspace}
\usepackage{ragged2e}
\usepackage{tabulary}
\usepackage{authblk}

\begin{document}

\title{Retrieval-Augmented Generation for Large Language Models: A Survey}

\author[a]{Yunfan Gao}
\author[b]{Yun Xiong}
\author[b]{Xinyu Gao}
\author[b]{Kangxiang Jia}
\author[b]{Jinliu Pan}
\author[c]{Yuxi Bi}
\author[a]{Yi Dai}
\author[a]{Jiawei Sun}
\author[c]{Meng Wang}
\author[a,c]{Haofen Wang \thanks{Corresponding Author.Email:\url{haofen.wang@tongji.edu.cn}}}

\affil[a]{Shanghai Research Institute for Intelligent Autonomous Systems, Tongji University}
\affil[b]{Shanghai Key Laboratory of Data Science, School of Computer Science, Fudan University}
\affil[c]{College of Design and Innovation, Tongji University}

\maketitle

\begin{abstract}
Large Language Models (LLMs) showcase impressive capabilities but encounter challenges like hallucination, outdated knowledge, and non-transparent, untraceable reasoning processes. Retrieval-Augmented Generation (RAG) has emerged as a promising solution by incorporating knowledge from external databases. This enhances the accuracy and credibility of the generation, particularly for knowledge-intensive tasks, and allows for continuous knowledge updates and integration of domain-specific information. RAG synergistically merges LLMs' intrinsic knowledge with the vast, dynamic repositories of external databases. This comprehensive review paper offers a detailed examination of the progression of RAG paradigms, encompassing the Naive RAG, the Advanced RAG, and the Modular RAG. It meticulously scrutinizes the tripartite foundation of RAG frameworks, which includes the retrieval, the generation and the augmentation techniques. The paper highlights the state-of-the-art technologies embedded in each of these critical components, providing a profound understanding of the advancements in RAG systems. Furthermore, this paper introduces up-to-date evaluation framework and benchmark. At the end, this article delineates the challenges currently faced and points out  prospective avenues for research and development~\footnote{Resources are available at \url{https://github.com/Tongji-KGLLM/RAG-Survey} }.

\end{abstract}


\section{Introduction}

%
\IEEEPARstart{L}{arge} language models (LLMs) have achieved remarkable success, though they still face significant limitations, especially in domain-specific or knowledge-intensive tasks~\cite{longtail}, notably producing ``hallucinations"~\cite{hallucination} when handling queries beyond their training data or requiring current information. To overcome challenges, Retrieval-Augmented Generation (RAG) enhances LLMs by retrieving relevant document chunks from external knowledge base through semantic similarity calculation. By referencing external knowledge, RAG effectively reduces the problem of generating factually incorrect content. Its integration into LLMs has resulted in widespread adoption, establishing RAG as a key technology in advancing chatbots and enhancing the suitability of LLMs for real-world applications.

\begin{figure*}
    \centering
    \includegraphics[width=0.85\linewidth]{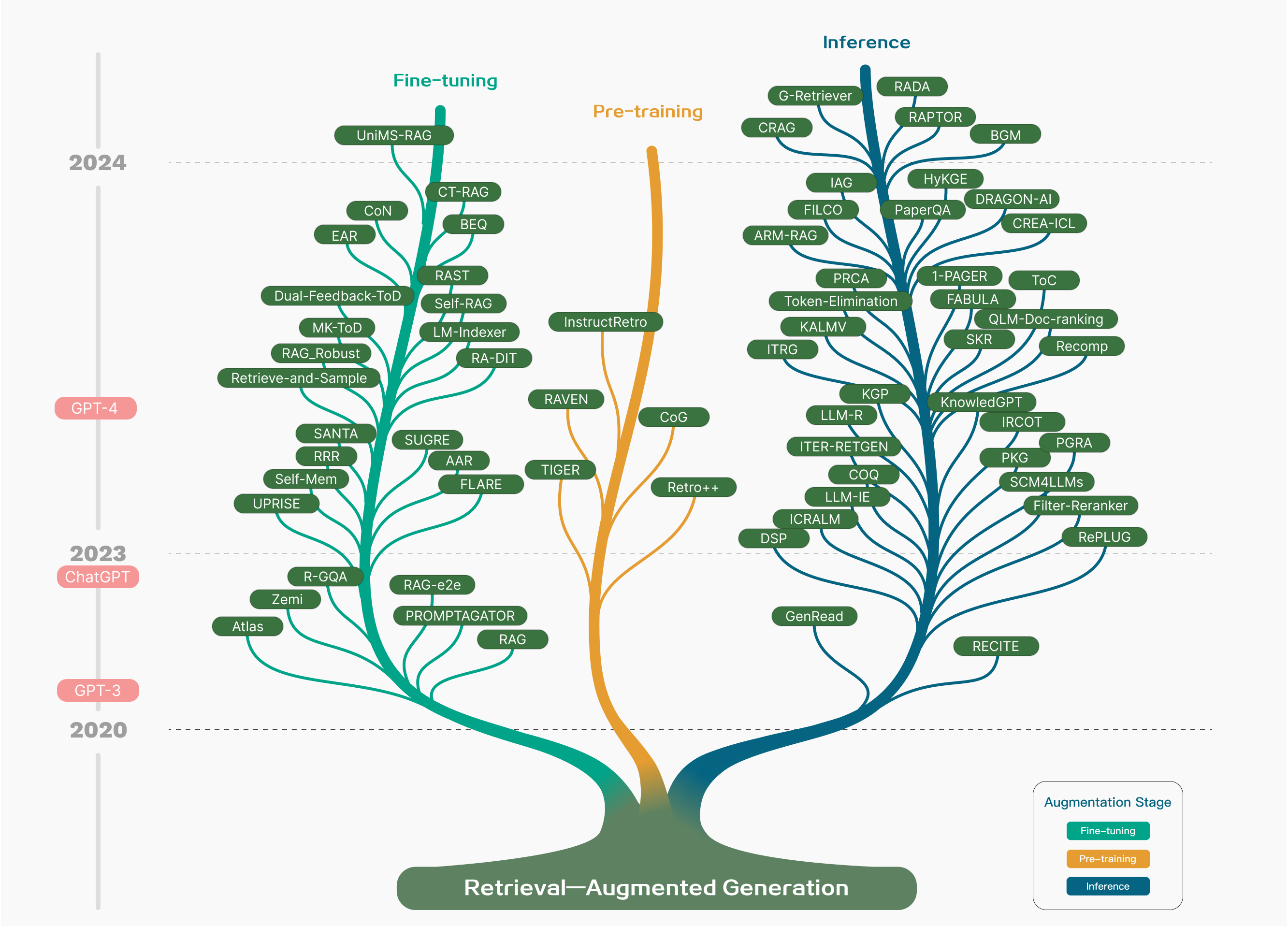}
    \caption{Technology tree of RAG research. The stages of involving RAG mainly include pre-training, fine-tuning, and inference. With the emergence of LLMs, research on RAG initially focused on leveraging the powerful in context learning abilities of LLMs, primarily concentrating on the inference stage. Subsequent research has delved deeper, gradually integrating more with the fine-tuning of LLMs. Researchers have also been exploring ways to enhance language models in the pre-training stage through retrieval-augmented techniques.}
    \label{fig:rag_tech_tree}
\end{figure*}


RAG technology has rapidly developed in recent years, and the technology tree summarizing related research is shown in Figure~\ref{fig:rag_tech_tree}. The development trajectory of RAG in the era of large models exhibits several distinct stage characteristics. Initially, RAG's inception coincided with the rise of the Transformer architecture, focusing on enhancing language models by incorporating additional knowledge through Pre-Training Models (PTM). This early stage was characterized by foundational work aimed at refining pre-training techniques\cite{REALM,RAG,Retro}.The subsequent arrival of ChatGPT~\cite{chatgpt} marked a pivotal moment, with LLM demonstrating powerful  in context learning (ICL) capabilities. RAG research shifted towards providing better information for LLMs to answer more complex and knowledge-intensive tasks during the inference stage, leading to rapid development in RAG studies. As research progressed, the enhancement of RAG was no longer limited to the inference stage but began to incorporate more with LLM fine-tuning techniques.

The burgeoning field of RAG has experienced swift growth, yet it has not been accompanied by a systematic synthesis that could clarify its broader trajectory. This survey endeavors to fill this gap by mapping out the RAG process and charting its evolution and anticipated future paths, with a focus on the integration of RAG within LLMs. This paper considers both technical paradigms and research methods, summarizing three main research paradigms from over 100 RAG studies, and analyzing key technologies in the core stages of ``Retrieval," ``Generation," and ``Augmentation." On the other hand, current research tends to focus more on methods, lacking analysis and summarization of how to evaluate RAG. This paper comprehensively reviews the downstream tasks, datasets, benchmarks, and evaluation methods applicable to RAG. Overall, this paper sets out to meticulously compile and categorize the foundational technical concepts, historical progression, and the spectrum of RAG methodologies and applications that have emerged post-LLMs. It is designed to equip readers and professionals with a detailed and structured understanding of both large models and RAG. It aims to illuminate the evolution of retrieval augmentation techniques, assess the strengths and weaknesses of various approaches in their respective contexts, and speculate on upcoming trends and innovations.

Our contributions are as follows:
\begin{itemize}
\item In this survey, we present a thorough and systematic review of the state-of-the-art RAG methods, delineating its evolution through paradigms including naive RAG, advanced RAG, and modular RAG. This review contextualizes the broader scope of RAG research within the landscape of LLMs.

\item We identify and discuss the central technologies integral to the RAG process, specifically focusing on the aspects of ``Retrieval", ``Generation'' and ``Augmentation", and delve into their synergies, elucidating how these components intricately collaborate to form a cohesive and effective RAG framework.

\item We have summarized the current assessment methods of RAG, covering 26 tasks, nearly 50 datasets, outlining the evaluation objectives and metrics, as well as the current evaluation benchmarks and tools. Additionally, we anticipate future directions for RAG, emphasizing potential enhancements to tackle current challenges.

\end{itemize}

The paper unfolds as follows: Section~\ref{sec:overview}  introduces the main concept and current paradigms of RAG. The following three sections explore core components—``Retrieval'', ``Generation" and ``Augmentation", respectively.
Section~\ref{sec:retrieval} focuses on optimization methods in retrieval,including indexing, query and embedding optimization.
Section~\ref{sec:generation} concentrates on post-retrieval process and LLM fine-tuning in generation.
Section~\ref{sec:augmentation} analyzes the three augmentation processes.
Section~\ref{sec:evaluation} focuses on RAG's downstream tasks and evaluation system. Section~\ref{sec:prospects}  mainly discusses the challenges that RAG currently faces and its future development directions. At last, the paper concludes in Section~\ref{sec:conclusion}.
\section{Overview of RAG }\label{sec:overview}

\begin{figure*}[tb]
    \centering
    \includegraphics[width=0.85\linewidth]{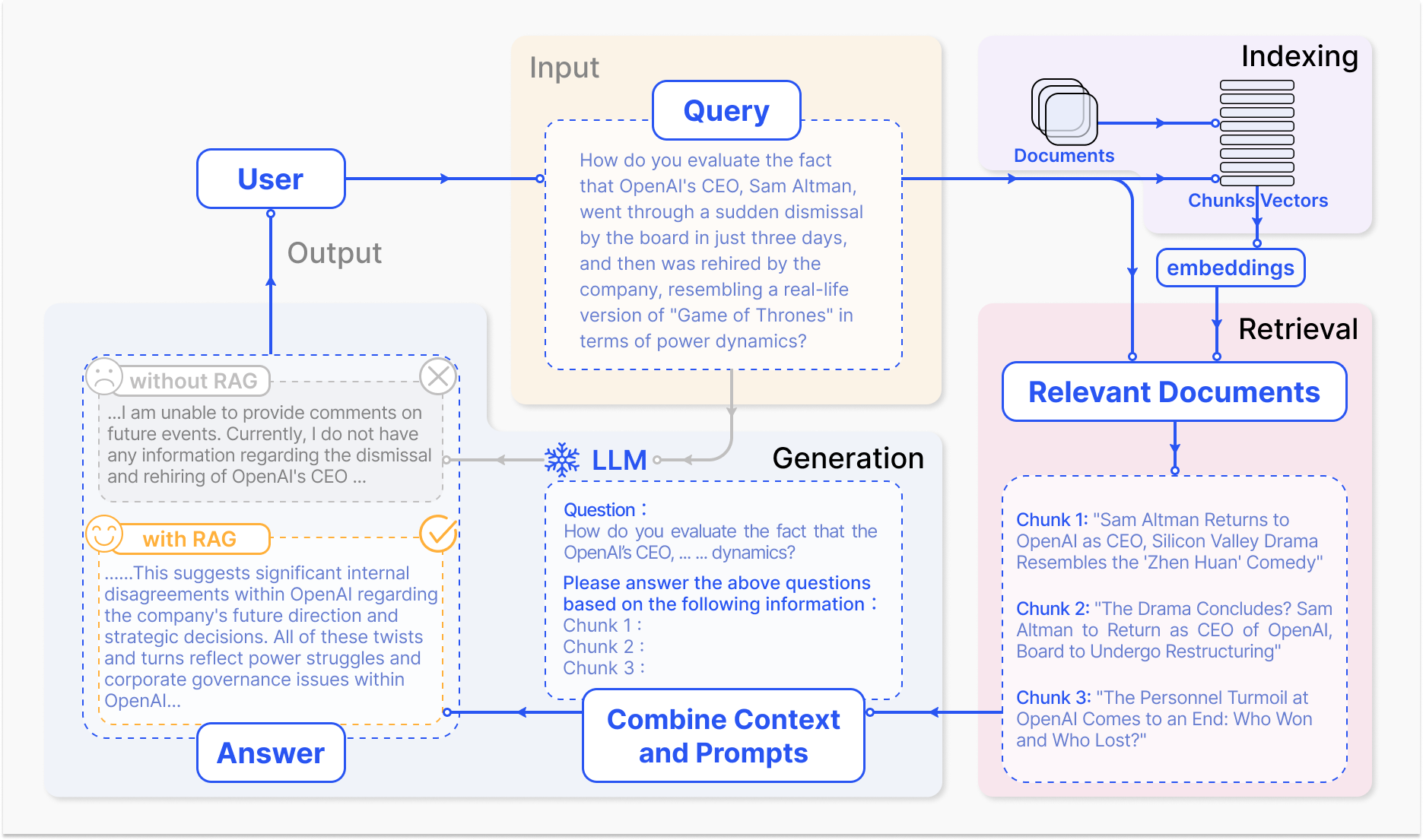}
    \caption{A representative instance of the RAG process applied to question answering. It mainly consists of 3 steps. 1) Indexing. Documents are split into chunks, encoded into vectors, and stored in a vector database. 2) Retrieval. Retrieve the Top k chunks most relevant to the question based on semantic similarity. 3) Generation. Input the original question and the retrieved chunks together into LLM to generate the final answer.}
    \label{fig:rag_case}
\end{figure*}

A typical application of RAG is illustrated in Figure~\ref{fig:rag_case}.  Here, a user poses a question to ChatGPT about a recent, widely discussed news. Given ChatGPT's reliance on pre-training data, it initially lacks the capacity to provide updates on recent developments. RAG bridges this information gap by sourcing and incorporating knowledge  from external databases. In this case, it gathers relevant news articles related to the user's query. These articles, combined with the original question, form a comprehensive prompt that empowers LLMs to generate a well-informed answer.  


\begin{figure*}
    \centering
    \includegraphics[scale=0.25]{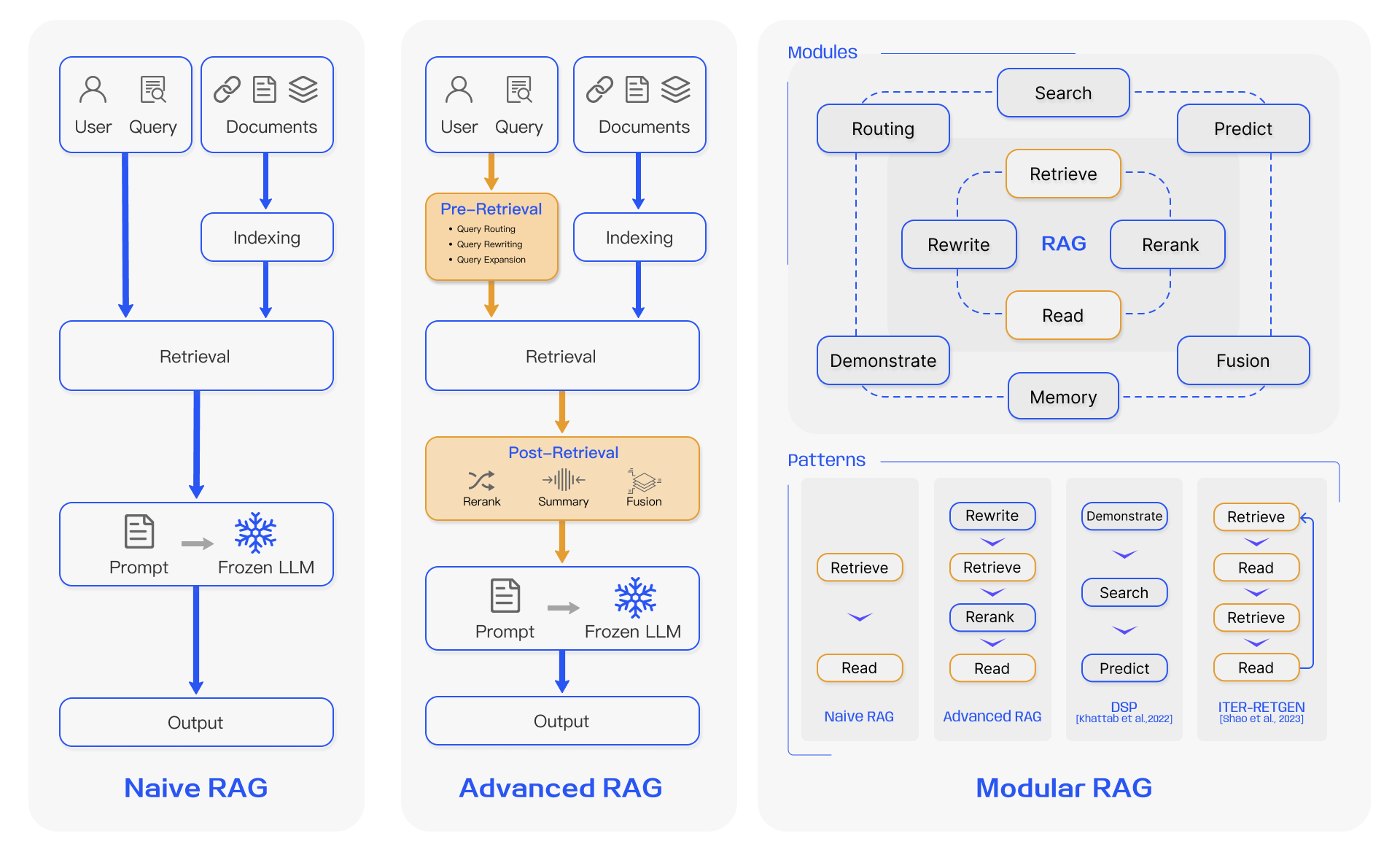}
    \caption{Comparison between the three paradigms of RAG. (Left) Naive RAG  mainly consists of three parts: indexing, retrieval and generation. (Middle) Advanced RAG proposes multiple optimization strategies around pre-retrieval and post-retrieval, with a process similar to the Naive RAG, still following a chain-like structure. (Right) Modular RAG inherits and develops from the previous paradigm, showcasing greater flexibility overall. This is evident in the introduction of multiple specific functional modules and the replacement of existing modules. The overall process is not limited to sequential retrieval and generation; it includes methods such as iterative and adaptive retrieval.}
    \label{fig:RAG_comp}
\end{figure*}

The RAG research paradigm is continuously evolving, and we categorize it into three stages: Naive RAG, Advanced RAG, and Modular RAG, as showed in Figure~\ref{fig:RAG_comp}. Despite RAG method are cost-effective and surpass the performance of the native LLM, they also exhibit several limitations. The development of Advanced RAG and Modular RAG is a response to these specific shortcomings in Naive RAG.
 
\subsection{Naive RAG}
The Naive RAG research paradigm represents the earliest methodology, which gained prominence shortly after the widespread adoption of ChatGPT. The Naive RAG follows a traditional process that includes indexing, retrieval, and generation, which is also characterized as a ``Retrieve-Read'' framework~\cite{RRR}.

\emph{Indexing} starts with the cleaning and extraction of raw data in diverse formats like PDF, HTML, Word, and Markdown, which is  then converted into a uniform plain text format. To accommodate the context limitations of language models,  text is segmented into smaller, digestible chunks. Chunks are then encoded into vector representations using an embedding model and stored in vector database. This step is crucial for enabling efficient similarity searches in the subsequent retrieval phase.

\emph{Retrieval}. Upon receipt of a user query,  the RAG system employs the same encoding model utilized during the indexing phase to transform the query into a vector representation. It then computes the similarity scores between the query vector and the vector of  chunks within the indexed corpus. The system prioritizes and retrieves the top K chunks that demonstrate the greatest similarity to the query. These chunks are subsequently used as the expanded context in prompt.

\emph{Generation}. The posed query and selected documents are synthesized into a coherent prompt to which a large  language model is tasked with formulating a response. The model's approach to answering may vary depending on task-specific criteria, allowing it to either draw upon its inherent parametric knowledge or restrict its responses to the information contained within the provided documents. In cases of ongoing dialogues, any existing conversational history can be integrated into the prompt, enabling the model to engage in multi-turn dialogue interactions effectively.

However, Naive RAG encounters notable drawbacks:

\emph{Retrieval Challenges}. The retrieval phase often struggles with precision and recall, leading to the selection of misaligned or irrelevant chunks, and the missing of crucial information.

\emph{Generation Difficulties}. In generating responses, the model may face the issue of hallucination, where it produces content not supported by the retrieved context. This phase can also suffer from irrelevance, toxicity, or bias in the outputs, detracting from the quality and reliability of the responses.

\emph{Augmentation Hurdles}. Integrating retrieved information with the different task can be challenging, sometimes resulting in disjointed or incoherent outputs. The process may also encounter redundancy when similar information is retrieved from multiple sources, leading to repetitive responses. Determining the significance and relevance of various passages and ensuring stylistic and tonal consistency add further complexity. Facing complex issues, a single retrieval based on the original query may not suffice to acquire adequate context information.

Moreover, there's a concern that generation models might overly rely on augmented information, leading to outputs that simply echo retrieved content without adding insightful or synthesized information.

\subsection{Advanced RAG}
Advanced RAG introduces specific improvements to overcome the limitations of Naive RAG. Focusing on enhancing retrieval quality, it employs pre-retrieval and post-retrieval strategies. To tackle the indexing issues, Advanced RAG refines its indexing techniques through the use of a sliding window approach, fine-grained segmentation, and the incorporation of metadata. Additionally, it incorporates several optimization methods to streamline the retrieval process\cite{Illustrat}. 

\emph{Pre-retrieval process}. In this stage, the primary focus is on optimizing the indexing structure and the original query. The goal of optimizing  indexing is to enhance the quality of the content being indexed. This involves strategies: enhancing data granularity, optimizing index structures, adding metadata, alignment optimization, and mixed retrieval. While the goal of query optimization is to make the user's original question clearer and more suitable for the retrieval task. Common methods include query rewriting query transformation, query expansion and other techniques~\cite{RRR,BEQUE,StepBack-prompt,HyDE}.

\emph{Post-Retrieval Process}. Once relevant context is retrieved, it's crucial to integrate it effectively with the query. The main methods in post-retrieval process include rerank chunks and context compressing.  Re-ranking the retrieved information to relocate the most relevant content to the edges of the prompt is a key strategy. This concept has been implemented in frameworks such as LlamaIndex\footnote{\url{https://www.llamaindex.ai}}, LangChain\footnote{\url{https://www.langchain.com/}}, and HayStack~\cite{LostInTheMiddleRanker}. Feeding all relevant documents directly into LLMs can lead to information overload, diluting the focus on key details with irrelevant content.To mitigate this, post-retrieval efforts concentrate on selecting the essential information, emphasizing critical sections, and shortening the context to be processed. 


    
\subsection{Modular RAG}

The modular RAG architecture advances beyond the former two RAG paradigms, offering enhanced adaptability and versatility. It incorporates diverse strategies for improving its components, such as adding a search module for similarity searches and refining the retriever through fine-tuning. Innovations like restructured RAG modules~\cite{GenRead} and rearranged RAG pipelines~\cite{ITER-RETGEN}  have been introduced to tackle specific challenges. The shift towards a modular RAG approach is becoming prevalent, supporting both sequential processing and integrated end-to-end training across its components. Despite its distinctiveness, Modular RAG builds upon the foundational principles of Advanced and Naive RAG, illustrating a progression and refinement within the RAG family. 

\subsubsection{New Modules}

The Modular RAG framework introduces additional specialized components to enhance retrieval and processing capabilities. The Search module adapts to specific scenarios, enabling direct searches across various data sources like search engines, databases, and knowledge graphs, using LLM-generated code and query languages~\cite{KnowledGPT}. RAG-Fusion addresses traditional search limitations by employing a multi-query strategy that expands user queries into diverse perspectives, utilizing parallel vector searches and intelligent re-ranking to uncover both explicit and transformative knowledge~\cite{fusion}. The Memory module leverages the LLM's memory to guide retrieval, creating an unbounded memory pool that aligns the text more closely with data distribution through iterative self-enhancement~\cite{selfmem,wang2022training}. Routing in the RAG system navigates through diverse data sources, selecting the optimal pathway for a query, whether it involves summarization, specific database searches, or merging different information streams~\cite{CREA-ICL}. The Predict module aims to reduce redundancy and noise by generating context directly through the LLM, ensuring relevance and accuracy~\cite{GenRead}. Lastly, the Task Adapter module tailors RAG to various downstream tasks, automating prompt retrieval for zero-shot inputs and creating task-specific retrievers through few-shot query generation~\cite{UPRISE,PROMPTAGATOR} .This comprehensive approach not only streamlines the retrieval process but also significantly improves the quality and relevance of the information retrieved, catering to a wide array of tasks and queries with enhanced precision and flexibility.

\subsubsection{New Patterns}
Modular RAG offers remarkable adaptability by allowing module substitution or reconfiguration to address specific challenges. This goes beyond the fixed structures of Naive and Advanced RAG, characterized by a simple ``Retrieve" and ``Read" mechanism. Moreover, Modular RAG expands this flexibility by integrating new modules or adjusting interaction flow among existing ones, enhancing its applicability across different tasks.

Innovations such as the Rewrite-Retrieve-Read~\cite{RRR}model leverage the LLM's capabilities to refine retrieval queries through a rewriting module and a LM-feedback mechanism to update rewriting model., improving task performance. Similarly, approaches like Generate-Read~\cite{GenRead} replace traditional retrieval with LLM-generated content, while Recite-Read~\cite{RECITE} emphasizes retrieval from model weights, enhancing the model's ability to handle knowledge-intensive tasks. Hybrid retrieval strategies integrate keyword, semantic, and vector searches to cater to diverse queries. Additionally, employing sub-queries and hypothetical document embeddings (HyDE)~\cite{HyDE} seeks to improve retrieval relevance by focusing on embedding similarities between generated answers and real documents.

Adjustments in module arrangement and interaction, such as the Demonstrate-Search-Predict (DSP)~\cite{DSP} framework and the iterative Retrieve-Read-Retrieve-Read flow of ITER-RETGEN~\cite{ITER-RETGEN}, showcase the dynamic use of module outputs to bolster another module's functionality, illustrating a sophisticated understanding of enhancing module synergy. The flexible orchestration of Modular RAG Flow showcases the benefits of adaptive retrieval through techniques such as FLARE~\cite{Flare} and Self-RAG~\cite{self-rag}. This approach transcends the fixed RAG retrieval process by evaluating the necessity of retrieval based on different scenarios. Another benefit of a flexible architecture is that the RAG system can more easily integrate with other technologies (such as fine-tuning or reinforcement learning)~\cite{BGM}. For example, this can involve fine-tuning the retriever for better retrieval results, fine-tuning the generator for more personalized outputs, or engaging in collaborative fine-tuning~\cite{RA-DIT}.

\subsection{RAG vs Fine-tuning}

\begin{figure*}[tb]
    \centering
    \includegraphics[width=0.8\linewidth]{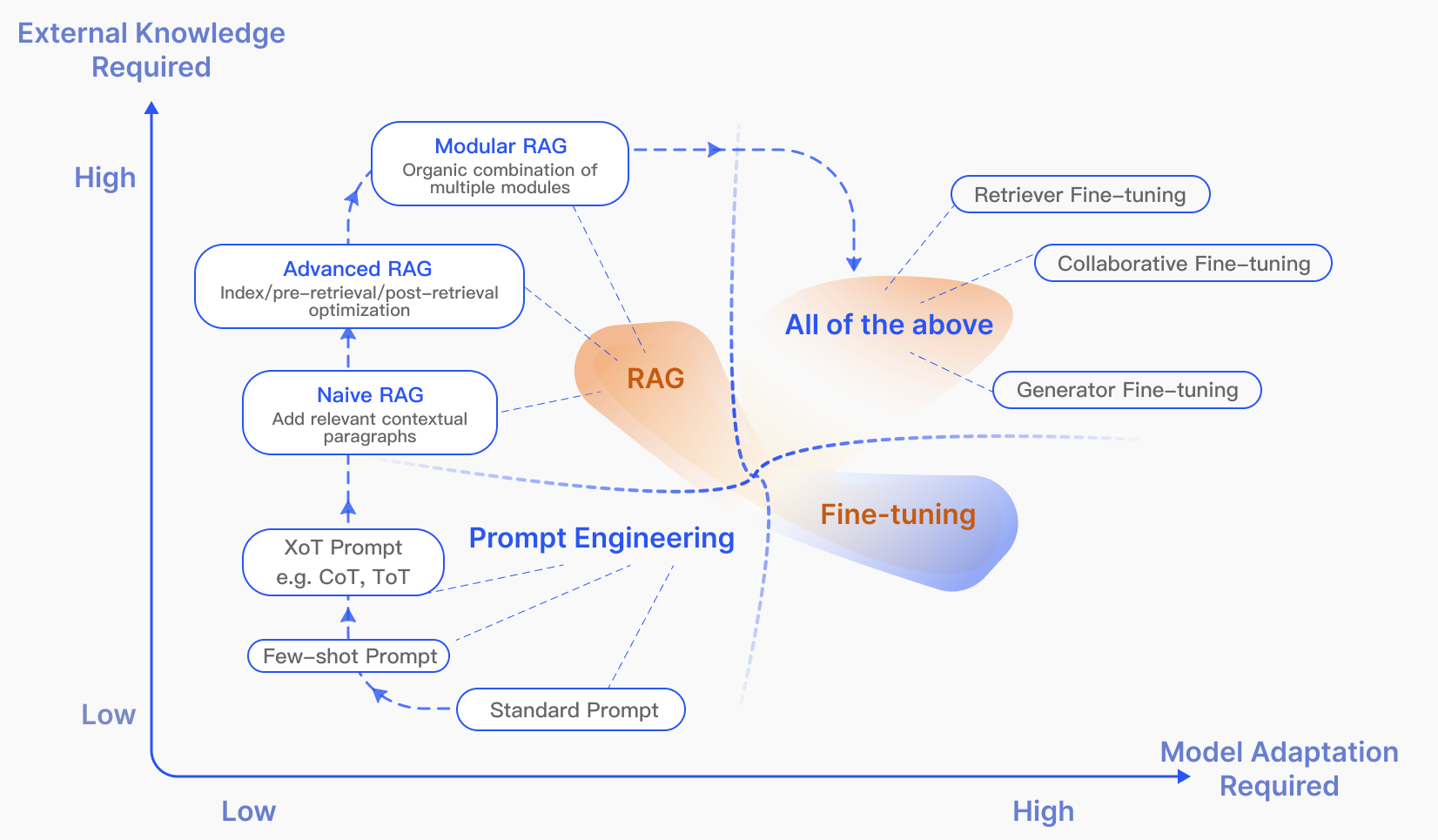}
    \caption{RAG compared with other model optimization methods in the aspects of ``External Knowledge Required" and ``Model Adaption Required". Prompt Engineering requires low modifications to the model and external knowledge, focusing on harnessing the capabilities of LLMs themselves. Fine-tuning, on the other hand, involves further training the model. In the early stages of RAG (Naive RAG), there is a low demand for model modifications. As research progresses, Modular RAG has become more integrated with fine-tuning techniques.}
    \label{fig:ragft}
\end{figure*}

The augmentation of LLMs has attracted considerable attention due to their growing prevalence. Among the optimization methods for LLMs, RAG is often compared with Fine-tuning (FT) and prompt engineering. Each method has distinct characteristics as illustrated in Figure~\ref{fig:ragft}. We used a quadrant chart to illustrate the differences among three methods in two dimensions: external knowledge requirements and model adaption requirements. Prompt engineering leverages a model's inherent capabilities with minimum necessity for external knowledge and model adaption. RAG can be likened to providing a model with a tailored textbook for information retrieval, ideal for precise information retrieval tasks. In contrast, FT is comparable to a student internalizing knowledge over time, suitable for scenarios requiring replication of specific structures, styles, or formats.  

RAG excels in dynamic environments by offering real-time knowledge updates and effective utilization of external knowledge sources with high interpretability. However, it comes with higher latency and ethical considerations regarding data retrieval. On the other hand, FT is more static, requiring retraining for updates but enabling deep customization of the model's behavior and style. It demands significant computational resources for dataset preparation and training, and while it can reduce hallucinations, it may face challenges with unfamiliar data.

In multiple evaluations of  their performance on various knowledge-intensive tasks across different topics,  \cite{FT-or-RAG} revealed that while unsupervised fine-tuning shows some improvement, RAG consistently outperforms it, for both existing knowledge encountered during training and entirely new knowledge. Additionally, it was found that LLMs struggle to learn new factual information through unsupervised fine-tuning. The choice between RAG and FT depends on the specific needs for data dynamics, customization, and computational capabilities in the application context. RAG and FT are not mutually exclusive and can complement each other, enhancing a model's capabilities at different levels.  In some instances, their combined use may lead to optimal performance. The optimization process involving RAG and FT may require multiple iterations to achieve satisfactory results.

\begin{table*}
\caption{Summary of RAG methods}
    \label{tab:RAG_summary}
    \centering
    \scalebox{0.8}{
    \begin{tabular}{cccccc} 
\toprule
\multicolumn{1}{c}{Method} & \multicolumn{1}{c}{Retrieval Source} & \multicolumn{1}{c}{\begin{tabular}[c]{@{}c@{}}Retrieval\\ Data Type\end{tabular}} & \multicolumn{1}{c}{\begin{tabular}[c]{@{}c@{}}Retrieval\\ Granularity\end{tabular}} & \multicolumn{1}{c}{\begin{tabular}[c]{@{}c@{}}Augmentation \\ Stage\end{tabular}} & \multicolumn{1}{c}{\begin{tabular}[c]{@{}c@{}}Retrieval \\ process\end{tabular}} \\

\midrule

CoG~\cite{COG}& Wikipedia& Text& Phrase& Pre-training&Iterative\\
DenseX~\cite{DenseX}& FactoidWiki& Text& Proposition& Inference&Once\\

EAR~\cite{EAR}& Dataset-base& Text& Sentence& Tuning&Once\\
UPRISE~\cite{UPRISE}& Dataset-base& Text& Sentence& Tuning&Once\\
RAST~\cite{RAST}& Dataset-base& Text& Sentence& Tuning&Once\\
Self-Mem~\cite{selfmem}& Dataset-base& Text& Sentence& Tuning&Iterative\\
FLARE~\cite{Flare}& Search Engine,Wikipedia& Text& Sentence& Tuning&Adaptive\\
PGRA~\cite{PGRA}& Wikipedia& Text& Sentence& Inference&Once\\
FILCO~\cite{FILCO}& Wikipedia& Text& Sentence& Inference&Once\\
RADA~\cite{RADA}& Dataset-base& Text& Sentence& Inference&Once\\

Filter-rerank~\cite{Filter-rerank}& Synthesized dataset& Text& Sentence& Inference&Once\\
R-GQA~\cite{R-GQA}& Dataset-base& Text& Sentence Pair& Tuning&Once\\
LLM-R~\cite{LLM-R}& Dataset-base& Text& Sentence Pair& Inference&Iterative\\

TIGER~\cite{TIGER}& Dataset-base& Text& Item-base& Pre-training&Once\\
LM-Indexer~\cite{LLM-Indexer}& Dataset-base& Text& Item-base& Tuning&Once\\
BEQUE~\cite{BEQUE}& Dataset-base& Text& Item-base& Tuning&Once\\
CT-RAG~\cite{CT-RAG}& Synthesized dataset& Text& Item-base& Tuning&Once\\

Atlas~\cite{Atlas}&   Wikipedia, Common Crawl&Text&  Chunk&  Pre-training&Iterative\\ 
RAVEN~\cite{RAVEN}& Wikipedia& Text& Chunk& Pre-training&Once\\
RETRO++~\cite{RETRO++}& Pre-training Corpus& Text& Chunk& Pre-training&Iterative\\
INSTRUCTRETRO~\cite{InstructRetro}&Pre-training corpus& Text& Chunk & Pre-training&Iterative\\
RRR~\cite{RRR}& Search Engine& Text& Chunk& Tuning&Once\\
RA-e2e~\cite{RAG-e2e}& Dataset-base& Text& Chunk& Tuning&Once\\
PROMPTAGATOR~\cite{PROMPTAGATOR}& BEIR&Text&  Chunk&  Tuning&Once\\
AAR~\cite{AAR}& MSMARCO,Wikipedia& Text& Chunk& Tuning&Once\\
RA-DIT~\cite{RA-DIT}& Common Crawl,Wikipedia& Text& Chunk& Tuning&Once\\
RAG-Robust~\cite{RAG-Robust}& Wikipedia& Text& Chunk& Tuning&Once\\
RA-Long-Form~\cite{RA-Long-Form} & Dataset-base& Text& Chunk& Tuning&Once\\
CoN~\cite{CoN}& Wikipedia& Text& Chunk& Tuning&Once\\
Self-RAG~\cite{self-rag}& Wikipedia& Text& Chunk& Tuning&Adaptive\\

BGM~\cite{BGM}& Wikipedia&Text & Chunk& Inference&Once\\
CoQ~\cite{CoQ}& Wikipedia& Text& Chunk& Inference&Iterative\\
Token-Elimination~\cite{token-elimination}& Wikipedia& Text& Chunk& Inference&Once\\
PaperQA~\cite{PaperQA}& Arxiv,Online Database,PubMed& Text& Chunk& Inference&Iterative\\
NoiseRAG~\cite{NoiseRAG}& FactoidWiki& Text& Chunk& Inference&Once\\
IAG~\cite{IAG}& Search Engine,Wikipedia& Text& Chunk& Inference&Once\\
NoMIRACL~\cite{NoMIRACL}& Wikipedia& Text& Chunk& Inference&Once\\
ToC~\cite{TOC}& Search Engine,Wikipedia& Text& Chunk& Inference&Recursive\\
SKR~\cite{SKR}& Dataset-base,Wikipedia& Text& Chunk& Inference&Adaptive\\
ITRG~\cite{ITRG}& Wikipedia& Text& Chunk& Inference&Iterative\\
RAG-LongContext~\cite{RAG-LongContext}& Dataset-base& Text& Chunk& Inference&Once\\
ITER-RETGEN~\cite{ITER-RETGEN}& Wikipedia& Text& Chunk& Inference&Iterative\\
IRCoT~\cite{IRCoT}& Wikipedia& Text& Chunk& Inference&Recursive\\
LLM-Knowledge-Boundary~\cite{LLM-boundary}& Wikipedia& Text& Chunk& Inference&Once\\
RAPTOR~\cite{RAPTOR}& Dataset-base&Text& Chunk& Inference&Recursive\\
RECITE~\cite{RECITE}&   LLMs&Text&  Chunk&  Inference&Once\\ 
ICRALM~\cite{ICRALM}& Pile,Wikipedia& Text& Chunk& Inference&Iterative\\

Retrieve-and-Sample~\cite{Retrieve-and-Sample}& Dataset-base& Text& Doc& Tuning&Once\\
Zemi~\cite{Zemi}& C4& Text& Doc& Tuning&Once\\
CRAG~\cite{CRAG}& Arxiv& Text& Doc & Inference&Once\\
1-PAGER~\cite{1-PAGER} & Wikipedia& Text& Doc& Inference&Iterative\\
PRCA~\cite{PRCA}& Dataset-base& Text& Doc& Inference&Once\\
QLM-Doc-ranking~\cite{QLM}& Dataset-base& Text& Doc& Inference&Once\\
Recomp~\cite{RECOMP}& Wikipedia& Text& Doc& Inference&Once\\
DSP~\cite{DSP}& Wikipedia& Text& Doc& Inference&Iterative\\
RePLUG~\cite{Replug}& Pile& Text& Doc& Inference&Once\\
ARM-RAG~\cite{ARM-RAG}& Dataset-base& Text& Doc& Inference&Iterative\\
GenRead~\cite{GenRead}&   LLMs&Text&  Doc&  Inference&Iterative\\

UniMS-RAG~\cite{UniMS-RAG}& Dataset-base& Text& Multi& Tuning&Once\\
CREA-ICL~\cite{CREA-ICL}& Dataset-base& Crosslingual,Text& Sentence& Inference&Once\\
PKG~\cite{PKG}& LLM& Tabular,Text& Chunk& Inference&Once\\
SANTA~\cite{SANTA}& Dataset-base& Code,Text & Item &Pre-training&Once\\
SURGE~\cite{sugre}&Freebase& KG& Sub-Graph& Tuning&Once\\
MK-ToD~\cite{MK-ToD}& Dataset-base& KG& Entity& Tuning&Once\\
Dual-Feedback-ToD~\cite{Dual-Feedback-TOD}& Dataset-base& KG& Entity Sequence& Tuning&Once\\
KnowledGPT~\cite{KnowledGPT}& Dataset-base& KG& Triplet& Inference&Muti-time\\
FABULA~\cite{Fabula}& Dataset-base,Graph& KG& Entity& Inference&Once\\
HyKGE~\cite{HyKGE}& CMeKG& KG& Entity& Inference&Once\\
KALMV~\cite{KALMV}& Wikipedia& KG&Triplet & Inference&Iterative\\
RoG~\cite{RoG}& Freebase& KG& Triplet& Inference&Iterative\\
G-Retriever~\cite{G-Retriever}& Dataset-base& TextGraph& Sub-Graph& Inference&Once\\

\bottomrule
\end{tabular}
}
\end{table*}

\section{Retrieval}\label{sec:retrieval}
In the context of RAG, it is crucial to efficiently retrieve relevant documents from the data source. There are several key issues involved, such as the retrieval source, retrieval granularity, pre-processing of the retrieval, and selection of the corresponding embedding model.

\subsection{Retrieval Source}
RAG relies on external knowledge to enhance LLMs, while the type of retrieval source and the granularity of retrieval units both affect the final generation results.

\subsubsection{Data Structure}
Initially, text is s the mainstream source of retrieval. Subsequently, the retrieval source expanded to include semi-structured data (PDF) and  structured data (Knowledge Graph, KG) for enhancement. In addition to retrieving from original external sources, there is also a growing trend in recent researches towards utilizing content generated by LLMs themselves for retrieval and enhancement purposes.

\emph{Unstructured Data}, such as text, is the most  widely used retrieval source, which are mainly gathered from corpus. For open-domain question-answering (ODQA) tasks, the primary retrieval sources are Wikipedia  Dump  with the current major versions including  HotpotQA~\footnote{\url{https://hotpotqa.github.io/wiki-readme.html}} (1st October , 2017), DPR\footnote{\url{https://github.com/facebookresearch/DPR}} (20 December, 2018). In addition to encyclopedic data, common unstructured data includes cross-lingual text~\cite{CREA-ICL} and domain-specific data (such as medical~\cite{CRAG}and legal domains~\cite{COG}). 

\emph{Semi-structured data}. typically refers to data that contains a combination of text and table information, such as PDF. Handling semi-structured data poses challenges for conventional RAG systems due to two main reasons. Firstly, text splitting processes may inadvertently separate tables, leading to data corruption during retrieval. Secondly, incorporating tables into the data can complicate semantic similarity searches. When dealing with semi-structured data, one approach involves leveraging the code capabilities of LLMs to execute Text-2-SQL queries on tables within databases, such as TableGPT~\cite{TableGPT}. Alternatively, tables can be transformed into text format for further analysis using text-based methods~\cite{PKG}. However, both of these methods are not optimal solutions, indicating substantial research opportunities in this area.

\emph{Structured data}, such as knowledge graphs (KGs)~\cite{iseeq} , which are typically verified and can  provide more precise information. KnowledGPT~\cite{KnowledGPT} generates KB search queries and stores knowledge in a personalized base, enhancing the RAG model's knowledge richness. In response to the limitations of LLMs in understanding and answering questions about textual graphs, G-Retriever~\cite{G-Retriever}  integrates Graph Neural Networks (GNNs), LLMs and RAG, enhancing graph comprehension and question-answering capabilities through soft prompting of the LLM, and employs the Prize-Collecting Steiner Tree (PCST) optimization problem for targeted graph retrieval. On the contrary, it requires additional effort to build, validate, and maintain structured databases. On the contrary, it requires additional effort to build, validate, and maintain structured databases.

\emph{LLMs-Generated Content.} Addressing the limitations of external auxiliary information in RAG, some research has focused on exploiting LLMs' internal knowledge. SKR~\cite{SKR} classifies questions as known or unknown, applying retrieval enhancement selectively. GenRead~\cite{GenRead} replaces the retriever with an LLM generator, finding that LLM-generated contexts often contain more accurate answers due to better alignment with the pre-training objectives of causal language modeling. Selfmem~\cite{selfmem} iteratively creates an unbounded memory pool with a retrieval-enhanced generator, using a memory selector to choose outputs that serve as dual problems to the original question, thus self-enhancing the generative model. These methodologies underscore the breadth of innovative data source utilization in RAG, striving to improve model performance and task effectiveness.

\subsubsection{Retrieval Granularity} 
Another important factor besides the data format of the retrieval source is the granularity of the retrieved data. Coarse-grained retrieval units theoretically can provide more relevant information for the problem, but they may also contain redundant content, which could distract the retriever and language models in downstream tasks~\cite{shi2023large,CoN}. On the other hand,  fine-grained retrieval unit granularity increases the burden of retrieval  and does not guarantee semantic integrity and meeting the required knowledge. Choosing the appropriate retrieval granularity during inference can be a simple and effective strategy to improve the retrieval and downstream task performance of dense retrievers.

In text, retrieval granularity ranges from fine to coarse, including Token, Phrase, Sentence, Proposition, Chunks, Document. Among them, DenseX~\cite{DenseX}proposed the concept of using propositions as retrieval units. Propositions are defined as atomic expressions in the text, each encapsulating a unique factual segment and presented in a concise, self-contained natural language format. This approach aims to enhance retrieval precision and relevance. On the Knowledge Graph (KG), retrieval granularity includes Entity, Triplet, and sub-Graph. The granularity of retrieval can also be adapted to downstream tasks, such as retrieving Item IDs~\cite{LLM-Indexer}in recommendation tasks and Sentence pairs~\cite{LLM-R}. Detailed information is illustrated in Table~\ref{tab:RAG_summary}.

\subsection{Indexing Optimization}
In the Indexing phase, documents will be processed, segmented, and transformed into Embeddings to be stored in a vector database. The quality of index construction determines whether the correct context can be obtained in the  retrieval phase.

\subsubsection{Chunking Strategy}
The most common method is to split the document into chunks on a fixed number of tokens (e.g., 100, 256, 512)~\cite{chunkszie}. Larger chunks can capture more context, but they also generate more noise, requiring longer processing time and higher costs. While smaller chunks may not fully convey the necessary context, they do have less noise. However, chunks leads to truncation within sentences, prompting the optimization of a recursive splits and sliding window  methods, enabling layered retrieval by merging globally related information across multiple retrieval processes~\cite{recursive}. Nevertheless, these approaches still cannot strike a balance between semantic completeness and  context length. Therefore, methods like Small2Big have been proposed, where sentences (small) are used as the retrieval unit, and the preceding and following sentences are provided as (big) context to LLMs~\cite{small2big}.

\subsubsection{Metadata Attachments}
Chunks can be enriched with metadata information such as page number, file name, author,category timestamp. Subsequently, retrieval can be filtered based on this metadata, limiting the scope of the retrieval. Assigning different weights to document timestamps during retrieval can achieve time-aware RAG, ensuring the freshness of knowledge and avoiding outdated information.

In addition to extracting metadata from the original documents, metadata can also be artificially constructed. For example, adding summaries of paragraph, as well as introducing hypothetical questions. This method is also known as Reverse HyDE. Specifically, using LLM to generate questions that can be answered by the document, then calculating the similarity between the original question and the hypothetical question during retrieval to reduce the semantic gap between the question and the answer.

\subsubsection{Structural Index}
One effective method for enhancing information retrieval is to establish a hierarchical structure for the documents. By constructing In structure, RAG system can expedite the retrieval and processing of pertinent data.

\emph{Hierarchical index structure}.  File are arranged in parent-child relationships, with chunks linked to them. Data summaries are stored at each node, aiding in the swift traversal of data and assisting the RAG system in determining which chunks to extract. This approach can also mitigate the illusion caused by block extraction issues.

\emph{Knowledge Graph index}. Utilize KG in constructing the hierarchical structure of documents contributes to maintaining consistency. It delineates the connections between different concepts and entities, markedly reducing the potential for illusions. Another advantage is the transformation of the information retrieval process into instructions that LLM can comprehend, thereby enhancing the accuracy of knowledge retrieval and enabling LLM to generate contextually coherent responses, thus improving the overall efficiency of the RAG system. To capture the logical relationship between document content and structure, KGP~\cite{KGP} proposed a method of building an index between multiple documents using KG. This KG consists of nodes (representing paragraphs or structures in the documents, such as pages and tables) and edges (indicating semantic/lexical similarity between paragraphs or relationships within the document structure), effectively addressing knowledge retrieval and reasoning problems in a multi-document environment.

\subsection{Query Optimization}

One of the primary challenges with Naive RAG is its direct reliance on the user’s original query as the basis for retrieval. Formulating a precise and clear question is difficult, and imprudent queries result in subpar retrieval effectiveness. Sometimes, the question itself is complex, and the language is not well-organized. Another difficulty lies in language complexity  ambiguity. Language models often struggle when dealing with specialized vocabulary or ambiguous abbreviations with multiple meanings. For instance, they may not discern whether “LLM” refers to \textit{large language model} or a \textit{Master of Laws} in a legal context.

\subsubsection{Query Expansion}
Expanding a single query into multiple queries enriches the content of the query, providing further context to address any lack of specific nuances, thereby ensuring the optimal relevance of the generated answers.

\emph{Multi-Query}. By employing prompt engineering to expand queries via LLMs, these queries can then be executed in parallel. The expansion of queries is not random, but rather meticulously designed. 

\emph{Sub-Query}. The process of sub-question planning represents the generation of the necessary sub-questions to contextualize and fully answer the original question when combined. This process of adding relevant context is, in principle, similar to query expansion. Specifically, a complex question can be decomposed into a series of simpler sub-questions using the least-to-most prompting method~\cite{least-to-most}.

\emph{Chain-of-Verification(CoVe)}. The expanded queries undergo validation by LLM to achieve the effect of reducing hallucinations. Validated expanded queries typically exhibit higher reliability~\cite{cove}.

\subsubsection{Query Transformation}
The core concept is to retrieve chunks based on a transformed query instead of the user’s original query.

\emph{Query Rewrite}.The original queries are not always optimal for LLM retrieval, especially in real-world scenarios. Therefore, we can prompt LLM to rewrite the queries. In addition to using LLM for query rewriting, specialized smaller language models, such as RRR (Rewrite-retrieve-read)~\cite{RRR}. The implementation of the query rewrite method in the Taobao, known as BEQUE~\cite{BEQUE} has notably enhanced recall effectiveness for long-tail queries, resulting in a rise in GMV.

Another query transformation method is to use prompt engineering to let LLM generate a query based on the original query for subsequent retrieval. HyDE~\cite{HyDE} construct  hypothetical documents  (assumed answers to the original query). It focuses on embedding similarity from answer to answer rather than seeking embedding similarity for the problem or query. Using the Step-back Prompting method~\cite{StepBack-prompt}, the original query is abstracted to generate a high-level concept question (step-back question). In the RAG system, both the step-back question and the original query are used for retrieval, and both the results are utilized as the basis for language model answer generation.

\subsubsection{Query Routing}
Based on varying queries, routing to distinct RAG pipeline,which is suitable for a versatile RAG system designed to accommodate diverse scenarios.

\emph{Metadata Router/ Filter}. The first step involves extracting keywords (entity) from the query, followed by filtering based on the keywords and metadata within the chunks to narrow down the search scope.

\emph{Semantic Router} is another method of routing involves leveraging the semantic information of the query. Specific apprach see Semantic Router~\footnote{\url{https://github.com/aurelio-labs/semantic-router}}. Certainly, a hybrid routing approach can also be employed, combining both semantic and metadata-based methods for enhanced query routing.

\subsection{Embedding}

In RAG, retrieval is achieved by calculating the similarity (e.g. cosine similarity) between the embeddings of the question and document chunks, where the semantic representation capability of embedding models plays a key role. This mainly includes a sparse encoder (BM25) and a dense retriever (BERT architecture Pre-training language models). Recent research has introduced prominent embedding models such as AngIE, Voyage, BGE,etc~\cite{AngIE,VOYAGE,BGE}, which are benefit from multi-task instruct tuning. Hugging Face’s MTEB leaderboard~\footnote{\url{https://huggingface.co/spaces/mteb/leaderboard}} evaluates embedding models across 8 tasks, covering 58 datasests. Additionally, C-MTEB  focuses on  Chinese capability, covering 6 tasks and 35 datasets. There is no one-size-fits-all answer to “which embedding model to use.” However,  some specific models are better suited for particular use cases.

\subsubsection{Mix/hybrid Retrieval }

Sparse and dense embedding approaches capture different relevance features and can benefit from each other by leveraging complementary relevance information. For instance, sparse retrieval models can be used to provide initial search results for training dense retrieval models. Additionally, pre-training language models (PLMs) can be utilized to learn term weights to enhance sparse retrieval. Specifically, it also demonstrates that sparse retrieval models can enhance the zero-shot retrieval capability of dense retrieval models and assist dense retrievers in handling queries containing rare entities, thereby improving robustness.

\subsubsection{Fine-tuning Embedding Model}
In instances where the context significantly deviates from pre-training corpus, particularly within highly specialized disciplines such as healthcare, legal practice, and other sectors replete with proprietary jargon, fine-tuning the embedding model on your own domain dataset becomes essential to mitigate such discrepancies.

In addition to supplementing domain knowledge, another purpose of fine-tuning is to align the retriever and generator, for example, using the results of LLM as the supervision signal for fine-tuning, known as LSR (LM-supervised Retriever). PROMPTAGATOR~\cite{PROMPTAGATOR} utilizes the LLM as a few-shot query generator to create task-specific retrievers, addressing challenges in supervised fine-tuning, particularly in data-scarce domains. Another approach, LLM-Embedder~\cite{LLM-Embedder}, exploits LLMs to generate reward signals across multiple downstream tasks. The retriever is fine-tuned with two types of supervised signals: hard labels for the dataset and soft rewards from the LLMs. This dual-signal approach fosters a more effective fine-tuning process, tailoring the embedding model to diverse downstream applications. REPLUG~\cite{Replug} utilizes a retriever and an LLM to calculate the probability distributions of the retrieved documents and then performs supervised training by computing the KL divergence. This straightforward and effective training method enhances the performance of the retrieval model by using an LM as the supervisory signal, eliminating the need for specific cross-attention mechanisms. Moreover, inspired by RLHF (Reinforcement Learning from Human Feedback), utilizing LM-based feedback to reinforce the retriever through reinforcement learning.

\subsection{Adapter}
Fine-tuning models may present challenges, such as integrating functionality through an API or addressing constraints arising from limited local computational resources. Consequently, some approaches opt to incorporate an external adapter to aid in alignment.

To optimize the multi-task capabilities of LLM, UPRISE~\cite{UPRISE} trained a lightweight prompt retriever that can automatically retrieve prompts from a pre-built prompt pool that are suitable for a given zero-shot task input. AAR (Augmentation-Adapted Retriver)~\cite{AAR} introduces a universal adapter designed to accommodate multiple downstream tasks. While  PRCA~\cite{PRCA} add a pluggable reward-driven contextual adapter to enhance performance on specific tasks. BGM~\cite{BGM} keeps the retriever and LLM fixed,and trains a bridge Seq2Seq model in between. The bridge model aims to transform the retrieved information into a format that LLMs can work with effectively, allowing it to not only rerank but also dynamically select passages for each query, and potentially employ more advanced strategies like repetition.  Furthermore, PKG introduces an innovative method for integrating knowledge into white-box models via directive fine-tuning~\cite{PKG}. In this approach, the retriever module is directly substituted to generate relevant documents according to a query. This method assists in addressing the difficulties encountered during the fine-tuning process and enhances model performance.



\section{Generation}\label{sec:generation}

After retrieval, it is not a good practice to directly input all the retrieved information to the LLM for answering questions. Following will introduce adjustments from two perspectives: adjusting the retrieved content and adjusting the LLM.

\subsection{Context Curation}

Redundant information can interfere with the final generation of LLM, and overly long contexts can also lead LLM to the ``Lost in the middle" problem~\cite{lostinthemiddle}. Like humans, LLM tends to only focus on the beginning and end of long texts, while forgetting the middle portion. Therefore, in the RAG system, we typically need to further process the retrieved content.

\subsubsection{Reranking}

Reranking fundamentally reorders document chunks to highlight the most pertinent results first, effectively reducing the overall document pool, severing a dual purpose in information retrieval, acting as both an enhancer and a filter, delivering refined inputs for more precise language model processing~\cite{QLM}. Reranking can be performed using rule-based methods that depend on predefined metrics like Diversity, Relevance, and MRR, or model-based approaches like Encoder-Decoder models from the BERT series (e.g., SpanBERT), specialized reranking models such as Cohere rerank or bge-raranker-large, and general large language models like GPT~\cite{LostInTheMiddleRanker,chatRec}.

\subsubsection{Context Selection/Compression}

A common misconception in the RAG process is the belief that retrieving as many relevant documents as possible and concatenating them to form a lengthy retrieval prompt is beneficial. However, excessive context can introduce more noise, diminishing the LLM’s perception of key information .

(Long) LLMLingua~\cite{LLMLingua,longllmlingua} utilize small language models (SLMs) such as GPT-2 Small or LLaMA-7B, to detect and remove  unimportant tokens, transforming it into a form that is challenging for humans to comprehend but well understood by LLMs. This approach presents a direct and practical method for prompt compression, eliminating the need for additional training of LLMs while balancing language integrity and compression ratio. PRCA tackled this issue by training an information extractor~\cite{PRCA}. Similarly, RECOMP adopts a comparable approach by training an information condenser using contrastive learning~\cite{RECOMP}. Each training data point consists of one positive sample and five negative samples, and the encoder undergoes training using contrastive loss throughout this process~\cite{DPR} .

In addition to compressing the context,  reducing the number of documents aslo helps improve the accuracy of the model's answers. Ma et al.~\cite{filter-ranker} propose the ``Filter-Reranker" paradigm, which combines the strengths of LLMs and SLMs. In this paradigm, SLMs serve as filters, while LLMs function as reordering agents. The research shows that instructing LLMs to rearrange challenging samples identified by SLMs leads to significant improvements in various Information Extraction (IE) tasks. Another straightforward and effective approach involves having the LLM evaluate the retrieved content before generating the final answer. This allows the LLM to filter out documents with poor relevance through LLM critique. For instance, in Chatlaw~\cite{chatlaw}, the LLM is prompted to self-suggestion on the referenced legal provisions to assess their relevance.

\subsection{LLM Fine-tuning}

Targeted fine-tuning based on the scenario and data characteristics  on LLMs can yield better results. This is also one of the greatest advantages of using on-premise LLMs. When LLMs lack data in a specific domain, additional knowledge can be provided to the LLM through fine-tuning. Huggingface’s fine-tuning data can also be used as an initial step. 

Another benefit of fine-tuning is the ability to adjust the model’s input and output. For example, it can enable LLM to adapt to specific data formats and generate responses in a particular style as instructed~\cite{R-GQA}. For retrieval tasks that engage with structured data, the SANTA framework~\cite{SANTA} implements a tripartite training regimen to effectively encapsulate both structural and semantic nuances. The initial phase focuses on the retriever, where contrastive learning is harnessed to refine the query and document embeddings.

Aligning LLM outputs with human or retriever preferences through reinforcement learning is a potential approach. For instance, manually annotating the final generated answers and then providing feedback through reinforcement learning. In addition to aligning with human preferences, it is also possible to align with the preferences of fine-tuned models and retrievers~\cite{Dual-Feedback-TOD}. When circumstances prevent access to powerful proprietary models or larger parameter open-source models, a simple and effective method is to distill the more powerful models(e.g. GPT-4). Fine-tuning of LLM can also be coordinated with fine-tuning of the retriever to align preferences. A typical approach, such as RA-DIT~\cite{RA-DIT}, aligns the scoring functions between Retriever and Generator using KL divergence.

\section{Augmentation process in RAG}\label{sec:augmentation}

\begin{figure*}
    \centering
    \includegraphics[width=1\linewidth,height=8cm]{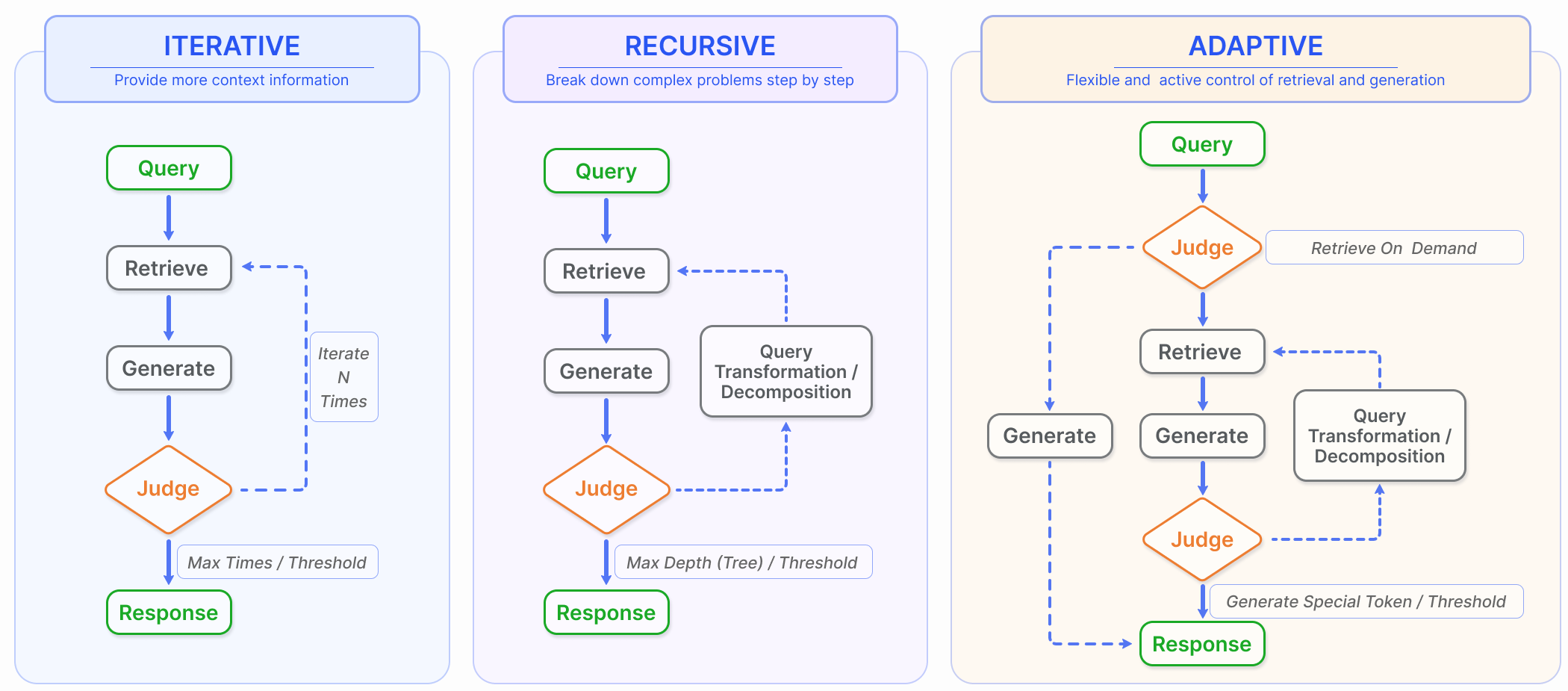}
    \caption{In addition to the most common once retrieval, RAG also includes three types of retrieval augmentation processes. (left) Iterative retrieval involves alternating between retrieval and generation, allowing for richer and more targeted context from the knowledge base at each step. (Middle) Recursive retrieval involves gradually refining the user query and breaking down the problem into sub-problems, then continuously solving complex problems through retrieval and generation. (Right) Adaptive retrieval focuses on enabling the RAG system to autonomously determine whether external knowledge retrieval is necessary and when to stop retrieval and generation, often utilizing LLM-generated special tokens for control.}
    \label{fig:aug_process}
\end{figure*}

In the domain of RAG, the standard practice often involves a singular (once) retrieval step followed by generation, which can lead to inefficiencies and sometimes is typically insufficient for complex problems demanding multi-step reasoning, as it provides a limited scope of information~\cite{irrelevantRobust}. Many studies have optimized the retrieval process in response to this issue, and we have summarised them in Figure~\ref{fig:aug_process}.

\subsection{Iterative Retrieval}

Iterative retrieval is a process where the knowledge base is repeatedly searched based on the initial query and the text generated so far, providing a more comprehensive knowledge base for LLMs. This approach has been shown to enhance the robustness of subsequent answer generation by offering additional contextual references through multiple retrieval iterations. However, it may be affected by semantic discontinuity and the accumulation of irrelevant information. ITER-RETGEN~\cite{ITER-RETGEN} employs a synergistic approach that leverages ``retrieval-enhanced generation'' alongside ``generation-enhanced retrieval'' for tasks that necessitate the reproduction of specific information. The model harnesses the content required to address the input task as a contextual basis for retrieving pertinent knowledge, which in turn facilitates the generation of improved responses in subsequent iterations.

\subsection{Recursive Retrieval}

Recursive retrieval is often used in information retrieval and NLP to improve the depth and relevance of search results. The process involves iteratively refining search queries based on the results obtained from previous searches. Recursive Retrieval aims to enhance the search experience by gradually converging on the most pertinent information through a feedback loop. IRCoT~\cite{IRCoT} uses chain-of-thought to guide the retrieval process and refines the CoT with the obtained retrieval results. ToC~\cite{TOC} creates a clarification tree that systematically optimizes the ambiguous parts in the Query. It can be particularly useful in complex search scenarios where the user's needs are not entirely clear from the outset or where the information sought is highly specialized or nuanced. The recursive nature of the process allows for continuous learning and adaptation to the user's requirements, often resulting in improved satisfaction with the search outcomes.

To address specific data scenarios, recursive retrieval and multi-hop retrieval techniques are utilized together. Recursive retrieval involves a structured index to process and retrieve data in a hierarchical manner, which may include summarizing sections of a document or lengthy PDF before performing a retrieval based on this summary. Subsequently, a secondary retrieval within the document refines the search, embodying the recursive nature of the process. In contrast, multi-hop retrieval is designed to delve deeper into graph-structured data sources, extracting interconnected information~\cite{CoK}.

\subsection{Adaptive Retrieval}

Adaptive retrieval methods, exemplified by Flare~\cite{Flare} and Self-RAG~\cite{self-rag}, refine the RAG framework by enabling LLMs to actively determine the optimal moments and content for retrieval, thus enhancing the efficiency and relevance of the information sourced.

These methods are part of a broader trend wherein LLMs employ active judgment in their operations, as seen in model agents like AutoGPT, Toolformer, and Graph-Toolformer~\cite{AutoGPT, Toolformer, Graph-Toolformer}. Graph-Toolformer, for instance, divides its retrieval process into distinct steps where LLMs proactively use retrievers, apply Self-Ask techniques, and employ few-shot prompts to initiate search queries. This proactive stance allows LLMs to decide when to search for necessary information, akin to how an agent utilizes tools.

WebGPT~\cite{WebGPT} integrates a reinforcement learning framework to train the GPT-3 model in autonomously using a search engine during text generation. It navigates this process using special tokens that facilitate actions such as search engine queries, browsing results, and citing references, thereby expanding GPT-3's capabilities through the use of external search engines. Flare automates timing retrieval by monitoring the confidence of the generation process, as indicated by the probability of generated terms~\cite{Flare}. When the probability falls below a certain threshold would activates the retrieval system to collect relevant information, thus optimizing the retrieval cycle. Self-RAG~\cite{self-rag} introduces ``reflection tokens'' that allow the model to introspect its outputs. These tokens come in two varieties: ``retrieve'' and ``critic''. The model autonomously decides when to activate retrieval, or alternatively, a predefined threshold may trigger the process. During retrieval, the generator conducts a fragment-level beam search across multiple paragraphs to derive the most coherent sequence. Critic scores are used to update the subdivision scores, with the flexibility to adjust these weights during inference, tailoring the model's behavior. Self-RAG's design obviates the need for additional classifiers or reliance on Natural Language Inference (NLI) models, thus streamlining the decision-making process for when to engage retrieval mechanisms and improving the model's autonomous judgment capabilities in generating accurate responses.



\section{Task and Evaluation}\label{sec:evaluation}
The rapid advancement and growing adoption of RAG in the field of  NLP have propelled the evaluation of RAG models to the forefront of research in the LLMs community. The primary objective of this evaluation is to comprehend and optimize the performance of RAG models across diverse application scenarios.This chapter will mainly introduce the main downstream tasks of RAG, datasets, and how to evaluate RAG systems.

\subsection{Downstream Task}
The core task of RAG remains Question Answering (QA), including traditional single-hop/multi-hop QA, multiple-choice, domain-specific QA as well as  long-form scenarios suitable for RAG. In addition to QA, RAG is continuously being expanded into multiple downstream tasks, such as Information Extraction (IE), dialogue generation, code search, etc. The main downstream tasks of RAG and their corresponding datasets are summarized in Table ~\ref{tab:Task}.

\begin{table*}[]
\caption{Downstream tasks and datasets of RAG\label{tab:Task}}
\centering
\begin{tabular}{@{}lllll@{}}
\toprule
Task                   & Sub Task                   & Dataset                   & Method                                                       &  \\ \midrule
QA &
  Single-hop &
  Natural Qustion(NQ)~\cite{NQ} &
  \begin{tabular}[c]{@{}l@{}}\cite{DenseX,BGM, KALMV, token-elimination,CoN,FILCO, Atlas, ICRALM,InstructRetro,ITRG}\\ \cite{LLM-boundary,LLM-Indexer,NoiseRAG,RA-DIT,RAG,RAG-Integration, RAVEN, REALM, RECITE, RECOMP} \\
  \cite{Replug,RETRO++, UPRISE}
  \end{tabular} &
   \\
 &
   &
  TriviaQA(TQA)~\cite{TQA} &
  \begin{tabular}[c]{@{}l@{}}\cite{DenseX,CoN,FILCO,GenRead,ICRALM,InstructRetro} \\ \cite{ITRG,LLM-boundary,RA-DIT,RAG,RAG-Integration}\\ \cite{RAVEN, RECITE,RECOMP,Replug,RETRO++,self-rag}\end{tabular} &
   \\
                       &                            & SQuAD~\cite{SQuAD}                   & \cite{DenseX,DSP,InstructRetro,PRCA,RAG-Integration, RAST, UPRISE}&  \\
                       &                            & Web Questions(WebQ)~\cite{webQA}       & \cite{DenseX,1-PAGER, CoN,GenRead,RAG,REALM}                        &  \\
                       &                            & PopQA~\cite{PopQA}                     & \cite{CRAG,RRR,self-rag}                                           &  \\
                       &                            & MS MARCO~\cite{MSMarco}                & \cite{token-elimination, LLM-Indexer, RAG}                           &  \\ \cmidrule(lr){2-4}
 &
  Multi-hop &
  HotpotQA~\cite{HotpotQA} &
  \begin{tabular}[c]{@{}l@{}}\cite{BGM,KALMV,CoQ,EAR, FILCO,AAR,DSP,IRCoT}\\ \cite{ITER-RETGEN,ITRG,KGP,LLM-boundary,PRCA,RA-DIT,RECITE,RECOMP,RRR}\end{tabular} &
   \\
                       &                            & 2WikiMultiHopQA~\cite{2WikiMultihop}           & \cite{Flare,IRCoT,ITER-RETGEN,ITRG,KGP,RAG-Robust}                  &  \\
                       &                            & MuSiQue~\cite{MuSiQue}                   & \cite{CoQ,IRCoT,ITER-RETGEN,KGP}                                   &  \\ \cmidrule(lr){2-4}
                       & Long-form QA               & ELI5~\cite{ELI5}                      & \cite{CoQ,FILCO,RA-DIT,RA-Long-Form,RAVEN}                        &  \\
                       &                            & NarrativeQA(NQA)~\cite{NarrativeQA}          & \cite{RAPTOR,ReadAgent,InstructRetro,RAG-LongContext}               &  \\
                       &                            & ASQA~\cite{ASQA}                   & \cite{Flare,TOC}                                                    &  \\
                       &                            & QMSum(QM)~\cite{QMSum}                & \cite{ReadAgent,RAG-LongContext}                                   &  \\ \cmidrule(lr){2-4}
                       & Domain QA                  & Qasper~\cite{Qasper}              & \cite{RAPTOR,RAG-LongContext}                                       &  \\
                       &                            & COVID-QA~\cite{COVID-QA}                  & \cite{RADA,RAG-e2e}                                                 &  \\
                       &                            & CMB~\cite{CMB},MMCU\_Medical~\cite{MMCU}         & \cite{HyKGE}                                                        &  \\ \cmidrule(lr){2-4}
                       & Multi-Choice QA            & QuALITY~\cite{QuALITY}                   & \cite{RAPTOR,RAG-LongContext}                                     &  \\
                       &                            & ARC~\cite{ARC}                       & \cite{CRAG,self-rag}                                                 &  \\
                       &                            & CommonsenseQA~\cite{CommonsenseQA}             & \cite{SKR,Zemi}                                                     &  \\ \cmidrule(lr){2-4}
                       & Graph QA                   & GraphQA~\cite{G-Retriever}                   & \cite{G-Retriever}                                                &  \\ \cmidrule(r){1-4}
Dialog                 & Dialog Generation          & Wizard of Wikipedia (WoW)~\cite{WoW} & \cite{FILCO,Atlas,GenRead,RA-DIT}                                   &  \\
                       & Personal Dialog            & KBP~\cite{KBP}                       & \cite{SAFARI, UniMS-RAG}                                             &  \\
                       &                            & DuleMon~\cite{DuleMon}                   & \cite{UniMS-RAG}                                                    &  \\
                       & Task-oriented Dialog           & CamRest~\cite{CamRest}              & \cite{Dual-Feedback-TOD,MK-ToD}                                     &  \\
                       & Recommendation             & Amazon(Toys,Sport,Beauty)~\cite{Amazon} & \cite{LLM-Indexer,TIGER}                                            &  \\ \cmidrule(r){1-4}
IE & Event Argument Extraction  & WikiEvent\cite{WikiEvent}                 & \cite{R-GQA,Atlas,GenRead,RA-DIT}                                   &  \\
                       &                            & RAMS~\cite{RAMS}                      & \cite{Filter-rerank,R-GQA}                                          &  \\
                       & Relation Extraction                         & T-REx~\cite{T-Rex},ZsRE~\cite{ZsRE}                & \cite{CoQ,RA-DIT}                                                   &  \\ \cmidrule(r){1-4}
Reasoning              & Commonsense Reasoning      & HellaSwag~\cite{HellaSwag}            & \cite{UPRISE,Zemi}                                                  &  \\
                       & CoT Reasoning              & CoT Reasoning~\cite{CoT-Reasoning}             & \cite{RA-DIT}                                  &  \\
                       & Complex Reasoning          & CSQA~\cite{CSQA}                     & \cite{IAG}                                                          &  \\ \cmidrule(r){1-4}
                      
Others                 & Language Understanding     & MMLU~\cite{mmlu}                     & \cite{AAR,Atlas, FT-or-RAG,RA-DIT,RAVEN,Replug,RRR}                 &  \\
                       & Language Modeling          & WikiText-103~\cite{WikiText-103}            & \cite{COG,ICRALM,RECOMP,Retro}                           &  \\
                       &                            & StrategyQA~\cite{StrategyQA}            & \cite{CoQ,Flare,IAG,ITER-RETGEN,RAG-Robust,SKR}                    &  \\
                        & Fact Checking/Verification & FEVER~\cite{FEVER}                     & \cite{CoN,FILCO,Atlas,GenRead,RA-DIT,RAG}                           &  \\
                       &                            & PubHealth~\cite{PubHealth}                 & \cite{self-rag,CRAG}                                           &  \\ 
                       & Text Generation            & Biography~\cite{WikiBio}                 & \cite{CRAG}                                             &  \\
                       & Text Summarization         & WikiASP~\cite{WikiASP}                   & \cite{Flare}                                                        &  \\
                       &                            & XSum~\cite{XSum}                      & \cite{selfmem}                                                    &  \\
                       & Text Classification        & VioLens~\cite{VioLens}                   & \cite{CREA-ICL}                                        &  \\
                       &                            & TREC~\cite{TREC}                      & \cite{PGRA}                                                         &  \\
                       & Sentiment                  & SST-2~\cite{SST-2}                   & \cite{PGRA,UPRISE,LLM-R}                                            &  \\
                       & Code Search                & CodeSearchNet~\cite{CodeSearchNet}             & \cite{SANTA}                                        &  \\
                       & Robustness Evaluation      & NoMIRACL~\cite{NoMIRACL}                  & \cite{NoMIRACL}                                   &  \\
                       & Math                       & GSM8K~\cite{GSM8K}                     & \cite{ARM-RAG}                                                      &  \\ 
                       & Machine Translation        & JRC-Acquis~\cite{JRC-Acquis}                & \cite{selfmem}                                                 & \\ \cmidrule(l){5-5} 
                       \midrule
\end{tabular}
\end{table*}

\subsection{Evaluation Target} 

Historically, RAG models assessments have centered on their execution in specific downstream tasks. These evaluations employ established metrics suitable to the tasks at hand. For instance, question answering evaluations might rely on EM and F1 scores~\cite{InstructRetro,Replug,ITRG,RRR}, whereas fact-checking tasks often hinge on Accuracy as the primary metric~\cite{RAG,Atlas,ITER-RETGEN}. BLEU and ROUGE metrics are also commonly used to evaluate answer quality~\cite{BGM,MK-ToD,token-elimination,RAST}. Tools like RALLE, designed for the automatic evaluation of RAG applications, similarly base their assessments on these task-specific metrics~\cite{RALLE}. Despite this, there is a notable paucity of research dedicated to evaluating the distinct characteristics of RAG models.The main evaluation objectives include:

\emph{Retrieval Quality}. Evaluating the retrieval quality is crucial for determining the effectiveness of the context sourced by the retriever component. Standard metrics from the domains of search engines, recommendation systems, and information retrieval systems are employed to measure the performance of the RAG retrieval module. Metrics such as Hit Rate, MRR, and NDCG are commonly utilized for this purpose~\cite{LlamaIndexTalk,DeepSet-Blog}.

\emph{Generation Quality}. The assessment of generation quality centers on the generator's capacity to synthesize coherent and relevant answers from the retrieved context. This evaluation can be categorized based on the content's objectives: unlabeled and labeled content. For unlabeled content, the evaluation encompasses the faithfulness, relevance, and non-harmfulness of the generated answers. In contrast, for labeled content, the focus is on the accuracy of the information produced by the model~\cite{LlamaIndexTalk}. Additionally, both retrieval and generation quality assessments can be conducted through manual or automatic evaluation methods~\cite{LlamaIndexTalk,COG,Databricks-RAG}.

\subsection{Evaluation Aspects}
Contemporary evaluation practices of RAG models emphasize three primary quality scores and four essential abilities, which collectively inform the evaluation of the two principal targets of the RAG model: retrieval and generation.

\subsubsection{Quality Scores}
Quality scores include context relevance, answer faithfulness, and answer relevance. These quality scores evaluate the efficiency of the RAG model from different perspectives in the process of information retrieval and generation~\cite{RAGAS,ARES,OpenAI-DevDayTalk}.

\emph{Context Relevance} evaluates the precision and specificity of the retrieved context, ensuring relevance and minimizing processing costs associated with extraneous content.

\emph{Answer Faithfulness} ensures that the generated answers remain true to the retrieved context, maintaining consistency and avoiding contradictions.

\emph{Answer Relevance} requires that the generated answers are directly pertinent to the posed questions, effectively addressing the core inquiry.

\subsubsection{Required Abilities}

RAG evaluation also encompasses four abilities indicative of its adaptability and efficiency: noise robustness, negative rejection, information integration, and counterfactual robustness~\cite{RGB,RECALL}. These abilities are critical for the model's performance under various challenges and complex scenarios, impacting the quality scores. 

\emph{Noise Robustness} appraises the model's capability to manage noise documents that are question-related but lack substantive information.

\emph{Negative Rejection} assesses the model's discernment in refraining from responding when the retrieved documents do not contain the necessary knowledge to answer a question.

\emph{Information Integration} evaluates the model's proficiency in synthesizing information from multiple documents to address complex questions.

 \emph{Counterfactual Robustness} tests the model's ability to recognize and disregard known inaccuracies within documents, even when instructed about potential misinformation. 

Context relevance and noise robustness are important for evaluating the quality of retrieval, while answer faithfulness, answer relevance, negative rejection, information integration, and counterfactual robustness are important for evaluating the quality of generation.

The specific metrics for each evaluation aspect are summarized in Table~\ref{tab:metrics-evaluation aspects}. It is essential to recognize that these metrics, derived from related work, are traditional measures and do not yet represent a mature or standardized approach for quantifying RAG evaluation aspects. Custom metrics tailored to the nuances of RAG models, though not included here, have also been developed in some evaluation studies.

\begin{table*}[htbp]
\centering
\caption{Summary of metrics applicable for evaluation aspects of RAG}
\begin{tabulary}{\linewidth}{@{}L*{7}{C}@{}}
\toprule
                & \thead{Context\\ Relevance} & \thead{Faithfulness} & \thead{Answer\\ Relevance} & \thead{Noise\\ Robustness} & \thead{Negative\\ Rejection} & \thead{Information\\ Integration} & \thead{Counterfactual\\ Robustness} \\ \midrule
Accuracy        & \checkmark                  & \checkmark           & \checkmark                & \checkmark                  & \checkmark                    & \checkmark                     & \checkmark                       \\
EM              &                             &                      &                           &                             & \checkmark                             &                                &                                  \\
Recall          & \checkmark                  &                      &                           &                             &                               &                                &                                  \\
Precision       &    \checkmark                         &           &                           &  \checkmark                            &                               &                                &                                  \\
R-Rate          &                             &                      &                 &                             &                               &                                & \checkmark                       \\
Cosine Similarity &                           &            & \checkmark                           &                             &                               &                                &                                  \\
Hit Rate        & \checkmark                  &                      &                           &                             &                               &                                &                                  \\
MRR             & \checkmark                  &                      &                           &                             &                               &                                &                                  \\
NDCG            & \checkmark                  &                      &                           &                             &                               &                                &     \\
BLEU              & \checkmark          &       \checkmark          &   \checkmark       &  &  &    & \\
ROUGE/ROUGE-L      &  \checkmark   & \checkmark & \checkmark  &  & & &

\\ \bottomrule
\end{tabulary}
\label{tab:metrics-evaluation aspects}
\end{table*}

\subsection{Evaluation Benchmarks and Tools}
A series of benchmark tests and tools have been proposed to facilitate the evaluation of RAG.These instruments furnish quantitative metrics that not only gauge RAG model performance but also enhance comprehension of the model's capabilities across various evaluation aspects. Prominent benchmarks such as RGB, RECALL and CRUD ~\cite{RGB,RECALL,CRUD} focus on appraising the essential abilities of RAG models. Concurrently, state-of-the-art automated tools like RAGAS~\cite{RAGAS}, ARES~\cite{ARES}, and TruLens\footnote{\url{https://www.trulens.org/trulens_eval/core_concepts_rag_triad/}} employ LLMs to adjudicate the quality scores. These tools and benchmarks collectively form a robust framework for the systematic evaluation of RAG models, as summarized in Table~\ref{tab:evaluation-frameworks}.

\begin{table*}[htbp]
\caption{Summary of evaluation frameworks}
\label{tab:evaluation-frameworks}
\centering
\begin{tabulary}{\textwidth}{@{}CCCCC@{}}
\toprule
\thead{\textbf{Evaluation Framework}} & \thead{\textbf{Evaluation Targets}} & \thead{\textbf{Evaluation Aspects}} & \thead{\textbf{Quantitative Metrics}} \\
\midrule
\makecell{RGB$^\dagger$} & \makecell{Retrieval Quality\\ Generation Quality} & \makecell{Noise Robustness\\ Negative Rejection\\ Information Integration\\ Counterfactual Robustness} & \makecell{Accuracy\\ EM\\ Accuracy\\ Accuracy} \\
\midrule
\makecell{RECALL$^\dagger$} & Generation Quality & Counterfactual Robustness & R-Rate (Reappearance Rate) \\
\midrule
\makecell{RAGAS$^\ddagger$} & \makecell{Retrieval Quality\\ Generation Quality} & \makecell{Context Relevance\\ Faithfulness\\ Answer Relevance} & \makecell{* \\ * \\ Cosine Similarity} \\
\midrule
\makecell{ARES$^\ddagger$} & \makecell{Retrieval Quality\\ Generation Quality} & \makecell{Context Relevance\\ Faithfulness\\ Answer Relevance} & \makecell{Accuracy\\ Accuracy\\ Accuracy} \\
\midrule
\makecell{TruLens$^\ddagger$} & \makecell{Retrieval Quality\\ Generation Quality} & \makecell{Context Relevance\\ Faithfulness\\ Answer Relevance} & \makecell{* \\ * \\ *} \\
\midrule
\makecell{CRUD$^\dagger$} & \makecell{Retrieval Quality\\ Generation Quality} & \makecell{Creative Generation\\ Knowledge-intensive QA\\ Error Correction \\Summarization} & \makecell{BLEU\\ROUGE-L\\BertScore\\RAGQuestEval  } \\
\bottomrule
\end{tabulary}
\par\bigskip
\textit{† represents a benchmark, and ‡ represents a tool. * denotes customized quantitative metrics, which deviate from traditional metrics. Readers are encouraged to consult pertinent literature for the specific quantification formulas associated with these metrics, as required.}
\end{table*}

\section{Discussion and Future Prospects}\label{sec:prospects}
Despite the considerable progress in RAG technology, several challenges persist that warrant in-depth research.This chapter will mainly introduce the current challenges and future research directions faced by RAG.

\subsection{RAG vs Long Context}
With the deepening of related research, the context of LLMs is continuously expanding~\cite{RAGLongcontext,packer2023memgpt,xiao2023efficient}. Presently, LLMs can effortlessly manage contexts exceeding 200,000 tokens~\footnote{\url{https://kimi.moonshot.cn}}. This capability signifies that long-document question answering, previously reliant on RAG, can now incorporate the entire document directly into the prompt. This has also sparked discussions on whether RAG is still necessary when LLMs are not constrained by context. In fact, RAG still plays an irreplaceable role. On one hand, providing LLMs with a large amount of context at once will significantly impact its inference speed, while chunked retrieval and on-demand input can significantly improve operational efficiency. On the other hand, RAG-based generation can quickly locate the original references for LLMs to help users verify the generated answers. The entire retrieval and reasoning process is observable, while generation solely relying on long context remains a black box. Conversely, the expansion of context provides new opportunities for the development of RAG, enabling it to address more complex problems and integrative or summary questions that require reading a large amount of material to answer~\cite{RA-Long-Form}. Developing new RAG methods in the context of super-long contexts is one of the future research trends.

\subsection{RAG Robustness}
The presence of noise or contradictory information during retrieval can detrimentally affect RAG's output quality. This situation is figuratively referred to as ``Misinformation can be worse than no information at all''. Improving RAG's resistance to such adversarial or counterfactual inputs is gaining research momentum and has become a key performance metric~\cite{CoN,RAG-Robust,KALMV}. Cuconasu et al.~\cite{NoiseRAG} analyze which type of documents should be retrieved, evaluate the relevance of the documents to the prompt, their position, and the number included in the context. The research findings reveal that including irrelevant documents can unexpectedly increase accuracy by over 30\%, contradicting the initial assumption of reduced quality. These results underscore the importance of developing specialized strategies to integrate retrieval with language generation models, highlighting the need for further research and exploration into the robustness of RAG.

\subsection{Hybrid Approaches }

Combining RAG with fine-tuning is emerging as a leading strategy. Determining the optimal integration of RAG and fine-tuning whether sequential, alternating, or through end-to-end joint training—and how to harness both parameterized and non-parameterized advantages are areas ripe for exploration~\cite{RA-DIT}. Another trend is to introduce SLMs with specific functionalities into RAG and fine-tuned by the results of RAG system. For example, CRAG~\cite{CRAG} trains a lightweight retrieval evaluator to assess the overall quality of the retrieved documents for a query and triggers different knowledge retrieval actions based on confidence levels. 

\subsection{Scaling laws of RAG }
End-to-end RAG models and pre-trained models based on RAG are still one of the focuses of current researchers~\cite{RAFT}.The parameters of these models are one of the key factors.While scaling laws~\cite{kaplan2020scaling} are established for LLMs, their applicability to RAG remains uncertain. Initial studies like RETRO++~\cite{RETRO++} have begun to address this, yet the parameter count in RAG models still lags behind that of LLMs. The possibility of an Inverse Scaling Law~\footnote{\url{https://github.com/inverse-scaling/prize}}, where smaller models outperform larger ones, is particularly intriguing and merits further investigation.

\subsection{Production-Ready RAG}
RAG's practicality and alignment with engineering requirements have facilitated its adoption. However, enhancing retrieval efficiency, improving document recall in large knowledge bases, and ensuring data security—such as preventing inadvertent disclosure of document sources or metadata by LLMs—are critical engineering challenges that remain to be addressed~\cite{retromaton}.

The development of the RAG ecosystem is greatly impacted by the progression of its technical stack. Key tools like LangChain and LLamaIndex have quickly gained popularity with the emergence of ChatGPT, providing extensive RAG-related APIs and becoming essential in the realm of LLMs.The emerging technology stack, while not as rich in features as LangChain and LLamaIndex, stands out through its specialized products. For example, Flowise AI prioritizes a low-code approach, allowing users to deploy AI applications, including RAG, through a user-friendly drag-and-drop interface. Other technologies like HayStack, Meltano, and Cohere Coral are also gaining attention for their unique contributions to the field.

In addition to AI-focused vendors, traditional software and cloud service providers are expanding their offerings to include RAG-centric services. Weaviate's Verba~\footnote{\url{https://github.com/weaviate/Verba}} is designed for personal assistant applications, while Amazon's Kendra~\footnote{\url{https://aws.amazon.com/cn/kendra/}} offers intelligent enterprise search services, enabling users to browse various content repositories using built-in connectors. In the development of RAG technology, there is a clear trend towards different specialization directions, such as: 1) Customization - tailoring RAG to meet specific requirements. 2) Simplification - making RAG easier to use to reduce the initial learning curve. 3) Specialization - optimizing RAG to better serve production environments.

The mutual growth of RAG models and their technology stacks is evident; technological advancements continuously establish new standards for existing infrastructure. In turn, enhancements to the technology stack drive the development of RAG capabilities. RAG toolkits are converging into a foundational technology stack, laying the groundwork for advanced enterprise applications. However, a fully integrated, comprehensive platform concept is still in the future, requiring further innovation and development.

\begin{figure*}[htbp]
    \centering
    \includegraphics[scale=0.22]{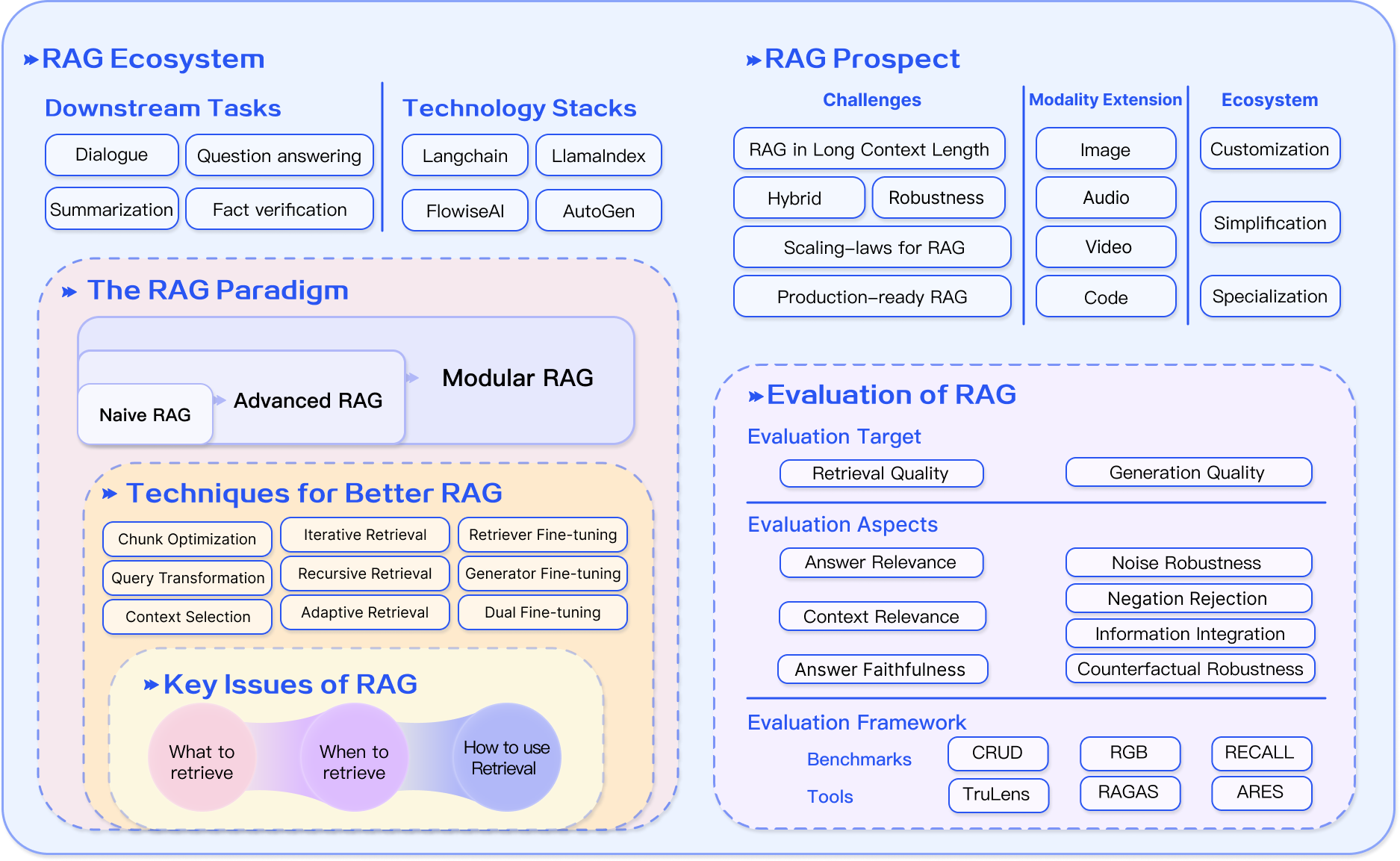}
    \caption{Summary of RAG ecosystem}
    \label{fig:rag_summary}
\end{figure*}

\subsection{Multi-modal RAG}
RAG has transcended its initial text-based question-answering confines, embracing a diverse array of modal data. This expansion has spawned innovative multimodal models that integrate RAG concepts across various domains:

\emph{Image}. RA-CM3~\cite{ramlm} stands as a pioneering multimodal model of both retrieving and generating text and images. BLIP-2~\cite{BLIP-2} leverages frozen image encoders alongside LLMs for efficient visual language pre-training, enabling zero-shot image-to-text conversions. The ``Visualize Before You Write'' method~\cite{VBR} employs image generation to steer the LM's text generation, showing promise in open-ended text generation tasks.

\emph{Audio and Video}. The GSS method retrieves and stitches together audio clips to convert machine-translated data into speech-translated data~\cite{GSS}. UEOP marks a significant advancement in end-to-end automatic speech recognition by incorporating external, offline strategies for voice-to-text conversion~\cite{UEOP}. Additionally, KNN-based attention fusion leverages audio embeddings and semantically related text embeddings to refine ASR, thereby accelerating domain adaptation. Vid2Seq augments language models with specialized temporal markers, facilitating the prediction of event boundaries and textual descriptions within a unified output sequence~\cite{Vid2Seq}.

\emph{Code}.  RBPS~\cite{RBPS} excels in small-scale learning tasks by retrieving code examples that align with developers' objectives through encoding and frequency analysis. This approach has demonstrated efficacy in tasks such as test assertion generation and program repair. For structured knowledge, the CoK method~\cite{CoK} first extracts facts pertinent to the input query from a knowledge graph, then integrates these facts as hints within the input, enhancing performance in knowledge graph question-answering tasks.

\section{Conclusion}\label{sec:conclusion}
The summary of this paper, as depicted in Figure~\ref{fig:rag_summary}, emphasizes RAG's significant advancement in enhancing the capabilities of LLMs by integrating parameterized knowledge from language models with extensive non-parameterized data from external knowledge bases. The survey showcases the evolution of RAG technologies and their application on many different tasks. The analysis outlines three developmental paradigms within the RAG framework: Naive, Advanced, and Modular RAG, each representing a progressive enhancement over its predecessors. RAG's technical integration with other AI methodologies, such as fine-tuning and reinforcement learning, has further expanded its capabilities. Despite the progress in RAG technology, there are research opportunities to improve its robustness and its ability to handle extended contexts. RAG's application scope is expanding into multimodal domains, adapting its principles to interpret and process diverse data forms like images, videos, and code. This expansion highlights RAG's significant practical implications for AI deployment, attracting interest from academic and industrial sectors. The growing ecosystem of RAG is evidenced by the rise in RAG-centric AI applications and the continuous development of supportive tools. As RAG's application landscape broadens, there is a need to refine evaluation methodologies to keep pace with its evolution. Ensuring accurate and representative performance assessments is crucial for fully capturing RAG's contributions to the AI research and development community.



\bibliographystyle{IEEEtran}

\bibliography{RAG}

\vfill

\end{document}